\documentclass{article}

\PassOptionsToPackage{numbers,sort&compress}{natbib}

    \usepackage[final]{neurips_2025}

\usepackage[utf8]{inputenc} 
\usepackage[T1]{fontenc}    
\usepackage{hyperref}       
\usepackage{url}            
\usepackage{booktabs}       
\usepackage{amsfonts}       
\usepackage{float}          
\usepackage{multirow}       
\usepackage{nicefrac}       
\usepackage{microtype}      
\usepackage{xcolor}         
\usepackage{makecell}    
\usepackage{graphicx}   
\usepackage{enumitem}
\usepackage{amsmath}

\usepackage[capitalize]{cleveref}

\title{Table2LaTeX-RL: High-Fidelity LaTeX Code Generation from Table Images via Reinforced Multimodal Language Models}

\author{
  Jun Ling\textsuperscript{1}   \\ \texttt{cs.lingjun@gmail.com} \\\And
  Yao Qi\textsuperscript{2} \And
  Tao Huang\textsuperscript{1} \\ \texttt{taohuang0313@gmail.com} \And
  Shibo Zhou\textsuperscript{2} \And
  Yanqin Huang\textsuperscript{2} \And
  Jiang Yang\textsuperscript{2} \And
  Ziqi Song\textsuperscript{2} \And
  Ying Zhou\textsuperscript{2} \And
  Yang Yang\textsuperscript{1} \And
  Heng Tao Shen\textsuperscript{1,3} \And
  Peng Wang\textsuperscript{1}\thanks{Corresponding author.} \\ \texttt{p.wang6@hotmail.com}\\\and
 \textsuperscript{1}School of Computer Science and Engineering, \\
  University of Electronic Science and Technology of China \\
  \textsuperscript{2}Research Center for Scientific Data Hub, Zhejiang Lab, Hangzhou, China \\
  \textsuperscript{3}School of Computer Science and Technology, Tongji University \\
}

\begin{document}
\maketitle

\begin{abstract}
In this work, we address the task of table image to LaTeX code generation, with the goal of automating the reconstruction of high-quality, publication-ready tables from visual inputs. A central challenge of this task lies in accurately handling complex tables—those with large sizes, deeply nested structures, and semantically rich or irregular cell content—where existing methods often fail. We begin with a comprehensive analysis, identifying key challenges and highlighting the limitations of current evaluation protocols. To overcome these issues, we propose a reinforced multimodal large language model (MLLM) framework, where a pre-trained MLLM is fine-tuned on a large-scale table-to-LaTeX dataset. To further improve generation quality, we introduce a dual-reward reinforcement learning strategy based on Group Relative Policy Optimization (GRPO). Unlike standard approaches that optimize purely over text outputs, our method incorporates both a structure-level reward on LaTeX code and a visual fidelity reward computed from rendered outputs, enabling direct optimization of the visual output quality.
We adopt a hybrid evaluation protocol combining TEDS-Structure and CW-SSIM, and show that our method achieves state-of-the-art performance, particularly on structurally complex tables, demonstrating the effectiveness and robustness of our approach. Code and dataset are available at https://github.com/newLLing/Table2LaTeX-RL.
\end{abstract}

\section{Introduction}
Tables are essential components of scientific and technical documents, providing a structured and concise format for presenting quantitative data, experimental results, and complex relationships. As document digitization becomes increasingly prevalent, the ability to automatically generate table code from images is critical for enabling content reuse and high-quality reproduction. However, most existing methods focus on generating HTML representations~\cite{zhong2020image,ye2021pingan,nassar2022tableformer, huang2023improving, shen2023divide} , which lack the structural expressiveness and typographic precision required for complex tables—especially those with nested headers, merged cells, or mathematical content. In contrast, LaTeX is the standard in scientific publishing, offering the flexibility and fidelity needed for professional-grade tables. Despite its practical importance, the task of directly generating LaTeX code from table images has received limited attention in prior work~\cite{Jiang2024LATTEIL, xia2025latexnet}.

In this work, we study the task of table image to LaTeX generation and provide a comprehensive analysis of its challenges. Through empirical observations, we find that the primary difficulty lies in handling complex tables, which are often large, deeply nested, and semantically rich—structures naturally suited to LaTeX but difficult for models to predict accurately. These challenges affect both the vision encoder, which must extract fine-grained visual and structural cues, and the language decoder, which must generate long, syntax-sensitive LaTeX sequences. Errors in either stage often lead to hallucinated, malformed output or even compilation errors. To enable finer-grained evaluation and better understand the current research gaps, we propose splitting the dataset into simple, medium, and complex subsets based on structural complexity.

To tackle these challenges, we leverage pre-trained multimodal large language models (MLLMs), which demonstrate strong capabilities in visual recognition, cross-modal reasoning, and LaTeX fluency. We fine-tune an MLLM on a large-scale image-to-LaTeX dataset harvested from scientific documents on arXiv. To further improve performance—particularly for complex tables—we introduce a dual-reward reinforcement learning strategy built on Group Relative Policy Optimization (GRPO)~\cite{Shao2024DeepSeekMathPT}, termed VSGRPO. While standard GRPO methods optimize text generation quality based solely on textual output, we go a step further: we render the generated LaTeX code into images and directly evaluate visual fidelity using CW-SSIM. This image-based reward complements a structure-level reward computed from the LaTeX source, allowing us to jointly optimize for both structural accuracy and rendered appearance. This novel visual-in-the-loop reinforcement design significantly enhances the model’s ability to produce faithful, high-fidelity LaTeX code for structurally rich and visually complex tables.

From an evaluation perspective, existing metrics are limited. TEDS~\cite{Zhong2019ImagebasedTR}, a widely used structure-based metric, lacks sensitivity to fine-grained errors and suffers from mismatches between HTML and LaTeX. On the other hand, rendered image comparison metrics focus on local visual similarity but ignore global structural correctness. To overcome this, we adopt a hybrid evaluation strategy that combines TEDS-Structure~\cite{Huang2023ImprovingTS}  for structural fidelity and CW-SSIM for robust visual similarity.

Under this framework, our method achieves state-of-the-art performance on the table image to LaTeX generation task, with particularly strong improvements on complex tables. This demonstrates the effectiveness of combining MLLM fine-tuning with targeted reinforcement learning for high-fidelity, publication-ready table generation.

\begin{itemize}
    \item We delve deep into the under-explored task of table image to LaTeX code generation, offering a comprehensive analysis of its core challenges—particularly for structurally complex tables—and introducing a complexity-based data split for fine-grained evaluation.
    \item We develop a reinforced MLLM framework, where a pre-trained MLLM is fine-tuned on a large-scale image-to-LaTeX dataset harvested from arXiv, effectively bridging visual input and LaTeX code generation.
    \item We propose VSGRPO, a novel dual-reward reinforcement learning strategy based on GRPO, which jointly optimizes structure-level accuracy and visual fidelity by incorporating both LaTeX-based and rendered-image-based feedback.
    \item We introduce a hybrid evaluation strategy combining TEDS-Structure and CW-SSIM to better reflect both structural and visual correctness. Extensive experiments demonstrate state-of-the-art performance of our approach, especially on complex tables.
    \end{itemize}

\section{Related Work}

\textbf{Table Structure Recognition.}\quad
Existing table recognition approaches fall into two main categories: detection-based and image-to-text–based methods. Detection-based methods first predict the physical structure—such as grid lines or cell bounding boxes—and then infer logical relationships. Grid-line-based approaches~\cite{schreiber2017deepdesrt,paliwal2019tablenet,tensmeyer2019deep,guo2022trust,zhang2022split,ma2023robust,lyu2023gridformer} segment tables along detected rows and columns and merge regions to reconstruct cells. Cell-bounding-box methods~\cite{raja2020table,liu2021show,wan2024omniparser,chi2019complicated} treat detected cells as graph nodes, using GNNs to infer row/column associations.

Image-to-text–based table structure recognition (TSR) decomposes the task into predicting structural layout and transcribing cell content, which are then fused into a full table representation. Encoder–decoder models~\cite{zhong2020image,ye2021pingan,nassar2022tableformer} generate structure tokens (e.g., HTML tags) and content separately. TableFormer~\cite{nassar2022tableformer} improves this with Transformer-based decoding and regression for bounding boxes. VAST~\cite{huang2023improving} frames coordinate prediction as sequence generation and adds a visual alignment loss. DRCC~\cite{shen2023divide} adopts a hybrid decoding scheme to reduce error accumulation.

Most detection and TSR methods target HTML outputs, which are not well-suited for LaTeX due to syntactic and semantic differences. Recently, end-to-end LaTeX generation approaches have emerged. LaTeXNet~\cite{xia2025latexnet} uses specialized submodules for equations, tables, and text. Nougat~\cite{blecher2023nougat} bypasses OCR entirely to generate LaTeX directly. LATTE~\cite{Jiang2024LATTEIL} introduces iterative refinement via localization and correction models. However, these methods do not explicitly address the combined challenges of large-scale layout and deeply nested LaTeX structures.

\textbf{Multimodal Large Language Models with Reinforced Fine-Tuning.}\quad
Pre-trained multimodal large language models (MLLMs) learn joint visual–text representations from large-scale image–text corpora, equipping them with strong capabilities in visual understanding and LaTeX code generation. While recent works such as Nougat~\cite{blecher2023nougat} and LATTE~\cite{Jiang2024LATTEIL} employ multimodal architectures, they largely underutilize pre-trained priors, relying instead on from-scratch training. 

To further improve performance, especially for complex tables, we apply reinforced fine-tuning using the  Group Relative Policy Optimization (GRPO) framework~\cite{Shao2024DeepSeekMathPT}. Compared to earlier reinforcement methods such as RLHF\cite{Sun2023AligningLM} and DPO\cite{Yu2023RLHFVTT}, GRPO eliminates the need for a value network and uses correctness-based rewards to guide learning with reduced computational overhead. Unlike prior works that apply reinforcement learning purely in the text domain~\cite{Huang2025VisionR1IR, Shen2025VLMR1AS}, our method designs task-specific reward signals: we compile the generated LaTeX into HTML for TEDS-Structure evaluation and into images for CW-SSIM computation, enabling joint optimization of both structural accuracy and rendered visual fidelity. This visual-in-the-loop RL approach is particularly effective for high-fidelity LaTeX generation on complex table structures.

\section{Insight into the Task}
We provide key insights into the task of table-to-LaTeX generation, focusing on two critical aspects: the challenge of handling complex tables and the limitations of current evaluation protocols.

One of the central challenges in this task lies in accurately processing complex tables, which serve as a meaningful indicator of a model's true capability. Complex tables are prevalent in modern documents, often used to convey large volumes of structured information compactly. However, their intricate layouts, large dimensions, and diverse content introduce significant difficulties for vision encoders—leading to higher computational costs, reduced performance, and increased inference latency. Despite their importance, complex tables are often underrepresented or overlooked in evaluation. 
To address this, we propose categorizing tables into three complexity levels—simple, medium, and complex—to enable a more realistic and fine-grained evaluation of model performance.

In addition to data-level challenges, the evaluation of LaTeX code generation remains underdeveloped, as shown in \Cref{case}. Existing metrics generally fall into two categories: \textbf{Text-based metrics}, such as TEDS~\cite{Zhong2019ImagebasedTR} and BLEU~\cite{Papineni2002BleuAM} , compare the predicted and ground-truth LaTeX code at the token level. However, they fail to account for the inherent syntactic ambiguity of LaTeX and ignore structural semantics, often penalizing semantically equivalent but syntactically different outputs. \textbf{Visual-based metrics}, such as CW-SSIM~\cite{CWSSIM}, evaluate similarity between rendered images of the generated and ground-truth tables. While useful for natural scenes, standard CW-SSIM is less effective on binary, high-contrast table images, where sharp edges and sparse textures dominate. Alternative pixel-level metrics, such as Edit (column-wise normalized edit distance) and Match (binary pixel-wise agreement), also fall short in capturing higher-level structural or semantic similarity. To address these limitations, we adopt a modified version of CW-SSIM, tuned for binary table images with high visual sparsity, to better assess rendering fidelity. However, since CW-SSIM primarily focuses on local visual similarity, we complement it with TEDS-Structure~\cite{Huang2023ImprovingTS} to evaluate the global structure and layout alignment.

\section{Method}


\subsection{Large-Scale Table2LaTeX Collection}

Due to the lack of publicly available large-scale datasets containing LaTeX table code, we propose a dataset construction pipeline. Specifically, we develop a web crawler to scrape the LaTeX source files of scientific papers from the open-access arXiv repository. We use regular expressions to extract LaTeX code corresponding to table environments. To ensure data quality, we further clean the extracted code by removing references, color settings, and other LaTeX control commands. Through this process, we collect a dataset comprising 1,209,986 table–LaTeX pairs.To classify table complexity, tables with 2 or more \verb|\multirow| or \verb|\multicolumn| commands and 100–160 cells are defined as medium tables, while those with over 160 cells are labeled complex tables. All others are considered simple.
 Within the training set, simple tables account for approximately 94\%, while medium and complex tables each represent about 3\% of the data.

\subsection{Supervised Fine-Tuning}

To enable general multimodal large language models (MLLMs) to acquire preliminary capability for handling the task of table-to-LaTeX generation, we design a second-stage supervised fine-tuning (SFT) process. During this process, we perform SFT using data collected in stage 1. The input consists of a table image and the prompt: \texttt{``Convert this table to LaTeX''}, while the ground-truth LaTeX code serves as the response. Thus, our dataset is structured as input-response pairs, formally expressed as: $\mathcal{D} \;=\; \bigl\{(\mathbf{x}^{(i)}, \mathbf{y}^{(i)})\bigr\}_{i=1}^N$
where each \(\mathbf{x}^{(i)}\) is an input and \(\mathbf{y}^{(i)}\) the corresponding target response. During training we optimize \(\theta\) to maximize the conditional likelihood of \(\mathbf{y}^{(i)}\) given \(\mathbf{x}^{(i)}\). Equivalently, we minimize the negative log‑likelihood over the dataset:
\begin{equation}
\label{eq2}
  \mathcal{L}_{\mathrm{SFT}}(\theta)
  = -\sum_{i=1}^N \log p_{\theta}\bigl(\mathbf{y}^{(i)} \mid \mathbf{x}^{(i)}\bigr).
\end{equation}

 However, as shown in~\Cref{tab:human_evaluation}, SFT alone is insufficient to fully unlock the model's potential. A key limitation stems from the widespread use of teacher forcing, where the model is trained to predict the next token given the prefix. Yet, LaTeX code is inherently ambiguous—different syntactic forms (e.g., control sequences) may yield identical visual outputs. This mismatch between training supervision and evaluation objectives leads to inefficient generalization, particularly for structurally complex tables.

\subsection{Reinforced Fine-Tuning via VSGRPO}

\begin{figure}[t]
    \small
    \centering
    \includegraphics[width=0.95\linewidth]{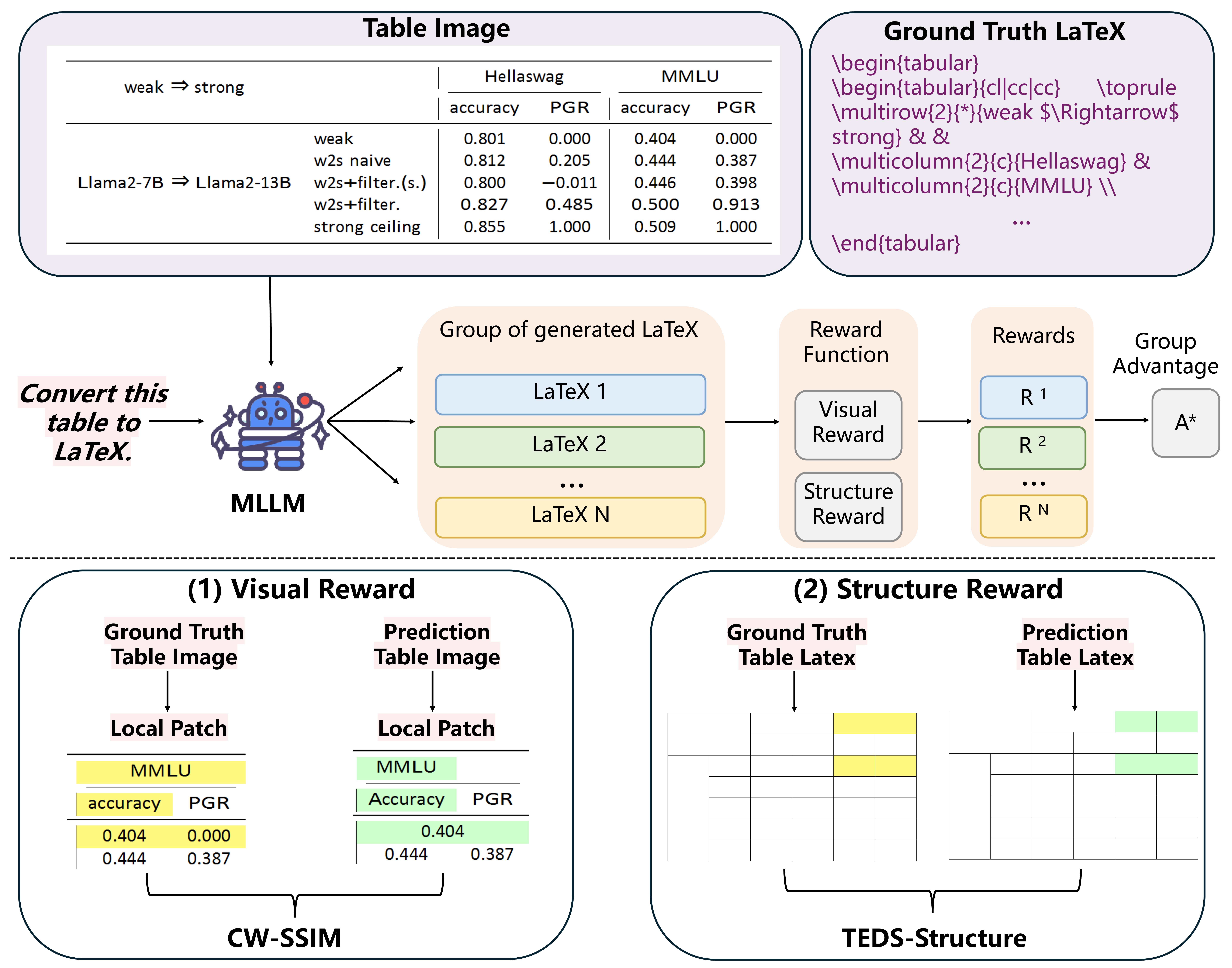}
    \caption{Demonstration of our proposed \textbf{VSGRPO} framework for table image to LaTeX code generation. The top section shows an example table image alongside its corresponding LaTeX code, representing the input-output pair used in training. The middle section illustrates the workflow of the VSGRPO framework. The bottom section highlights the dual-reward mechanism: a visual fidelity reward computed via CW-SSIM between the model-generated and ground-truth rendered images, and a structure-level reward based on TEDS-Structure computed from the table’s structural elements.}
    \label{fig:method}
      \vspace{-5pt}
\end{figure}

As analyzed above, the next-token prediction paradigm used in SFT is limited in its ability to model the semantic structure and syntactic dependencies embedded in long LaTeX sequences. Moreover, the SFT objective focuses solely on text-level alignment and completely ignores the visual similarity between the rendered LaTeX output and the original table image—despite visual appearance being a direct and critical indicator of generation quality. However, since LaTeX rendering is a non-differentiable operation, it cannot be directly incorporated into gradient-based supervised training.

To address these limitations, we propose a novel reinforced fine-tuning framework that introduces rendered image feedback as an explicit optimization signal. Drawing inspiration from  Group Relative Policy Optimization (GRPO)~\cite{Shao2024DeepSeekMathPT}, we extend its scope beyond standard textual quality assessment and design a dual-reward mechanism that jointly promotes structural accuracy and visual fidelity. While conventional GRPO-based methods focus solely on improving text generation quality, our framework leverages both the LaTeX code structure and its rendered appearance, offering a more task-aligned supervision signal, as shown in \Cref{fig:method}. We select 5,936 complex tables from the training dataset as the training set for VSGRPO, whose ground-truth LaTeX code contains fewer than 3,000 characters to balance complexity and computational feasibility.

\textbf{Visual Reward.}\quad We compile and convert the set of predicted table LaTeX code—generated by the model from a single table image input—by embedding them into LaTeX fragments with standard macro packages, producing a group of predicted table images. At the same time, we compile the ground-truth LaTeX code to obtain the ground-truth table image. If the compilation fails, the corresponding reward is set to 0. We then compute the CW-SSIM between the ground-truth table image and each predicted table image. If the CW-SSIM of a predicted image exceeds the predefined threshold, it receives a reward of 1; otherwise, the reward is 0.

To accommodate black-and-white table images, we adopt the following CW-SSIM calculation process: the CW-SSIM algorithm preprocesses two table images by converting them to grayscale, resizing them to uniform dimensions, and aligning their rows and columns. It then divides the images into 2×2 pixel blocks and applies a simplified Haar wavelet transform~\cite{schmidt1907theorie} to decompose each block into four sub-bands: cA (low-frequency approximation), cH (horizontal), cV (vertical), and cD (diagonal high-frequency details). For each sub-band, the algorithm calculates SSIM metrics optimized for monochrome tables, incorporating pixel-level means, variances, covariance, and stabilizing constants C\_1 and C\_2. Finally, it averages the SSIM scores from all four sub-bands to generate the comprehensive CW-SSIM metric.



\textbf{Structure Reward.}\quad We convert both the predicted table LaTeX code generated by the model and the ground-truth LaTeX into HTML in order to compute their TEDS-Structure. If the HTML conversion of a predicted table LaTeX fails, the reward is set to 0. For successfully converted predictions, if the TEDS-Structure similarity with the ground-truth exceeds a predefined threshold, the reward is set to 1; otherwise, it is 0.

TEDS-Structure computes the Minimum Tree Edit Distance between the two trees by applying unit-cost insertions, deletions, and structural-node substitutions to transform the predicted tree into the ground-truth tree. It then normalizes this distance by the larger of the two tree sizes and converts it into a similarity score.

During the RFT training process, we samples a group of generated output set $\{o_1,o_2,\cdots,o_N\}$ for each table image $q$ from policy model $\pi_{\theta_{old}}$. Then RFT maximizes the following objective and optimizes the model $\pi_{\theta}$. The specific formula is shown below:

\begin{equation}
\begin{aligned}
J_{\mathrm{RFT}}(\theta) 
&= \mathbb{E}_{q \sim P(Q),\, \{o_i\}_{i=1}^N \sim \pi_{\theta_{\mathrm{old}}}(O \mid q)} \\
&\Biggl[
  \frac{1}{N} \sum_{i=1}^N
  \min\!\Bigl(
    \frac{\pi_{\theta}(o_i \mid q)}{\pi_{\theta_{\mathrm{old}}}(o_i \mid q)}\,A_i,\,
    \mathrm{clip}\!\Bigl(
      \frac{\pi_{\theta}(o_i \mid q)}{\pi_{\theta_{\mathrm{old}}}(o_i \mid q)},
      1-\varepsilon,\,
      1+\varepsilon
    \Bigr)A_i
  \Bigr)
  -\,\beta\,D_{\mathrm{KL}}\bigl(\pi_{\theta}\,\big\|\,\pi_{\mathrm{ref}}\bigr)
\Biggr],
\end{aligned}
\label{eq3}
\end{equation}

where \(\varepsilon\) and \(\beta\) denote the clipping threshold in Proximal Policy Optimization (PPO) and the coefficient controlling the Kullback–Leibler (KL) divergence penalty term, respectively~\cite{shao2024deepseekmath,schulman2017proximal}. We set \(\varepsilon = 0.2\) and \(\beta = 0.02\) during training.

The advantage for the \(i\)-th sample is computed as

\begin{equation}\label{eq4}
A_i = \frac{r_i - \operatorname{mean}\bigl(\{r_1, r_2, \dots, r_N\}\bigr)}
               {\operatorname{std}\bigl(\{r_1, r_2, \dots, r_N\}\bigr)},
\end{equation}

where \(\{r_1, r_2, \dots, r_N\}\) denotes the set of group rewards. The KL divergence between the current policy \(\pi_\theta\) and the reference policy \(\pi_{\mathrm{ref}}\) for the observation–action pair \((q, o_i)\) is defined as
\begin{equation}\label{eq5}
D_{\mathrm{KL}}\bigl(\pi_\theta \,\|\, \pi_{\mathrm{ref}}\bigr)
= \frac{\pi_{\mathrm{ref}}(o_i \mid q)}{\pi_\theta(o_i \mid q)}
- \log\!\Bigl(\frac{\pi_{\mathrm{ref}}(o_i \mid q)}{\pi_\theta(o_i \mid q)}\Bigr)
- 1\,.
\end{equation}

\section{Experiments}

In this section, we present our experimental results. Specifically, \Cref{Experimental-setup} details the datasets, implementation settings, and evaluation metrics used in our study. \Cref{main_results} reports quantitative comparisons against state-of-the-art baselines. To further assess whether the generated table images align with human perception, we conduct a human evaluation, presented in \Cref{human_evaluation}. Finally, \Cref{ablation_study} provides an ablation analysis to highlight the key components and contributions of our proposed method.

\subsection{Experimental Setup}

\label{Experimental-setup}
In this section, we first describe the detailed construction process and composition of the training and testing datasets. We then present the implementation details for both the SFT and reinforced fine-tuning (RFT) phases.

\textbf{Training Dataset.}\quad
We collect LaTeX source code from arXiv papers published between October 2017 and April 2023, extracting a total of 1,209,986 table entries using regular expression matches between \verb|\begin{tabular}| and \verb|\end{tabular}|. After filtering out references, color commands, and other non-structural LaTeX elements, we classify the tables into three categories. This full dataset is used for SFT. 

\textbf{Testing Dataset.}\quad
We construct the testing dataset using the same processing pipeline as the training set. Specifically, we crawl 101,469 LaTeX table entries from arXiv papers published between January and November 2024, covering a diverse range of scientific domains. From this pool, we randomly sample 496 simple, 354 medium, and 361 complex tables to form the final testing set.

\textbf{Implementation Details.}\quad
We adopt full-parameter fine-tuning for all training phases. During SFT, all models are trained for one epoch with a maximum output length of 4096 tokens. For Nougat~\cite{blecher2023nougat}, training is conducted on 4 nodes (each with 8×A100 GPUs) using a batch size of 2. InternVL2-1B is trained on 4 nodes with a batch size of 4 and gradient accumulation steps set to 2. Qwen2.5-VL-3B~\cite{zheng2024llamafactory} is trained on 2 nodes, also with a batch size of 4 and gradient accumulation steps of 2.

For reinforced fine-tuning (RFT), InternVL2-1B adopts the VLM-R1 framework~\cite{shen2025vlm} and is trained on 2 nodes with 8 sampled generations per input (num\_gens = 8), a batch size of 8, and gradient accumulation steps of 2. Qwen2.5-VL-3B uses the ms-swift infrastructure~\cite{zhao2024swiftascalablelightweightinfrastructure}, trained on 2 nodes with 4 generations per input (num\_gens = 4), a batch size of 4, and the same gradient step setting. The reward thresholds are set to 0.6 for CW-SSIM and 0.9 for TEDS-Structure.

During testing, we use a maximum output length of 8192 tokens, a batch size of 1, and a temperature of 0. All testing is conducted within a \texttt{texlive-full} Docker environment to ensure LaTeX rendering fidelity. For metrics, we use Python-based implementations of CW-SSIM, TEDS-Structure, and TEDS for performance evaluation. The scores range from 0 to 1, and the exact formulas are shown in \Cref{metric}.



\subsection{Main Results}

\label{main_results}
We compare VSGRPO with various solutions across different categories. In the commercial and paid domain, we evaluate it against the most powerful system to date, Mathpix~\cite{mathpix2025}. To compare with current general-purpose multimodal large models, we include the closed-source GPT-4o~\cite{openai2025gpt4o}, as well as the open-source Qwen2.5-VL-72B~\cite{bai2025qwen2} and Intern2.5-VL-78B~\cite{chen2024expanding}. For specialized expert models, we compare against Nougat~\cite{blecher2023nougat}, a state-of-the-art open-source LaTeX generation system.

To more accurately evaluate the correctness of LaTeX generation, we assess model performance from two complementary perspectives: rendered image quality and LaTeX source fidelity. First, we evaluate the visual accuracy of the generated LaTeX by compiling it into table images. Two metrics are used: the compile ratio, which reflects the proportion of LaTeX outputs that can be successfully compiled using standard LaTeX packages, and CW-SSIM, which quantifies the visual similarity between the rendered output and the ground-truth image. These results are reported in \cref{tab:1}. Second, we assess the semantic and structural correctness of the LaTeX source code itself. To this end, we compute TEDS-Structure, which measures cell-level structural alignment, and TEDS, which additionally considers the tabular content. These metrics provide a deeper view into how well the generated code captures the underlying table semantics, and are summarized in \cref{tab:2}. To further evaluate the generalization ability of our method, we additionally test it on an external benchmark dataset introduced in~\cite{Jiang2024LATTEIL}, with the results shown in \cref{tab:3}.

\Cref{tab:1} shows that the CW-SSIM values of all models exhibit a decreasing trend as table complexity increases (from simple to complex). However, the proposed VSGRPO method achieves comprehensive improvements across two model families. Specifically, Intern2-VL-1B-VSGRPO sets a new record on the simple tables with a CW-SSIM of 0.8201, surpassing the previous best model by 0.049. Meanwhile, Qwen2.5-VL-3B-VSGRPO significantly outperforms baselines on the medium tables and complex tables, achieving CW-SSIM scores of 0.7236 (+0.1113) and 0.6145 (+0.0903), respectively. Furthermore, it attains a compile success rate of 0.9917 on the complex tables, exceeding Mathpix's 0.9889. These results demonstrate that the proposed VSGRPO strategy effectively enhances the robustness of complex tables reconstruction while maintaining high LaTeX compilability through visual- and structure-guided optimization.

\begin{table}[htbp]
  \vspace{-5pt}
\centering
\caption{Model performance on CW-SSIM and compile ratio across three table complexity levels.}
\label{tab:1}
\resizebox{1\textwidth}{!}{
\begin{tabular}{lcc|cc|cc}
\toprule
\multirow{2}{*}{Models} 
  & \multicolumn{2}{c}{Simple} 
  & \multicolumn{2}{c}{Medium} 
  & \multicolumn{2}{c}{Complex} \\ 
\cmidrule(lr){2-3} \cmidrule(lr){4-5} \cmidrule(lr){6-7}
  & CW-SSIM       & Compile ratio 
  & CW-SSIM       & Compile ratio 
  & CW-SSIM       & Compile ratio \\ 
\midrule
\multicolumn{7}{l}{\textit{Commercial Tools}} \\
Mathpix~\cite{mathpix2025}            & 0.6884 & \textbf{1.0000} & 0.5647 & 0.9943 & 0.4862 & 0.9889 \\
\midrule
\multicolumn{7}{l}{\textit{General VLMs}} \\
GPT4o~\cite{openai2025gpt4o}             & 0.6792 & 0.9918 & 0.5612 & \textbf{0.9972} & 0.4747 & 0.9917 \\
Qwen2.5-VL-72B~\cite{bai2025qwen2}    & 0.7077 & 0.9858 & 0.6009 & 0.9887 & 0.5112 & 0.9335 \\
Intern2.5-VL-78B~\cite{chen2024expanding}  & 0.7814 & 0.9959 & 0.6123 & 0.9773 & 0.5242 & 0.4515 \\
\midrule
\multicolumn{7}{l}{\textit{Expert VLMs}} \\
Nougat~\cite{blecher2023nougat}            & 0.7401 & 0.7617 & 0.5505 & 0.1813 & 0.4699 & 0.3352 \\
\midrule
\multicolumn{7}{l}{\textit{Our Results}} \\
Intern2-VL-1B-VSGRPO    & \textbf{0.8201} & 0.9939 & 0.7185 & 0.9830 & 0.5899 & 0.9640 \\
Qwen2.5-VL-3B-VSGRPO    & 0.8186 & 0.9980 & \textbf{0.7236} & 0.9943 & \textbf{0.6145} & \textbf{0.9917} \\
\bottomrule
\end{tabular}
}
  \vspace{-5pt}
\end{table}

\Cref{tab:2} presents results for TEDS and TEDS-Structure metrics. The trend of TEDS scores largely mirrors that of TEDS-Structure, although the absolute values are consistently lower due to TEDS additionally accounting for cell content alignment. 
The commercial tool Mathpix demonstrates relatively stable performance across table types, achieving its highest TEDS-Structure score on medium-complexity tables (0.8965). In the general-purpose VLM category, Qwen2.5-VL-72B shows consistently strong structural performance, with the highest TEDS-Structure score on simple tables (0.9400). However, it exhibits a gradual performance decline as complexity increases—TEDS drops from 0.8720 (simple) to 0.8090 (medium) and 0.7448 (complex). By contrast, other large-scale models such as Intern2.5-VL-78B experience a sharp drop on complex tables (TEDS: 0.3379), and the expert model Nougat collapses almost entirely (TEDS: 0.0424), revealing severe limitations in both structural and content-level generalization.
In contrast, our proposed Qwen2.5-VL-3B-VSGRPO achieves consistently superior results across all levels of table complexity. Despite its compact size (3B parameters), it outperforms significantly larger models, reaching a TEDS score of 0.8673 on complex tables—0.1225 higher than the next-best model—and achieving a TEDS-Structure score of 0.9218, the first to surpass the 0.9 threshold on complex tables. These results underscore the effectiveness of our dual-reward optimization strategy, which integrates structural and visual supervision to enable robust, high-fidelity LaTeX code generation, especially for structurally rich and visually complex tables.


\begin{table}[htbp]
  \vspace{-5pt}
\centering
\caption{Performance of different models on TEDS and TEDS-Structure across three table complexity levels.}
\label{tab:2}
\resizebox{1\textwidth}{!}{
\begin{tabular}{lcc|cc|cc}
\toprule
\multirow{2}{*}{Models} 
  & \multicolumn{2}{c}{Simple}              
  & \multicolumn{2}{c}{Medium}              
  & \multicolumn{2}{c}{Complex}             \\ 
\cmidrule(lr){2-3}\cmidrule(lr){4-5}\cmidrule(lr){6-7}
    & TEDS & TEDS-Structure
    & TEDS & TEDS-Structure
    & TEDS & TEDS-Structure \\ 
\midrule
\multicolumn{7}{l}{\textit{Commercial Tools}}                                                                                                           \\
Mathpix~\cite{mathpix2025}                
  & 0.7804     & 0.8701   
  & 0.8044     & 0.8965   
  & 0.7176     & 0.8100  \\ 
\midrule
\multicolumn{7}{l}{\textit{General VLMs}}                                                                                                             \\
GPT4o~\cite{openai2025gpt4o}                  
  & 0.8259     & 0.9117   
  & 0.6986     & 0.8451   
  & 0.5865     & 0.7745  \\
Qwen2.5-VL-72B~\cite{bai2025qwen2}          
  & 0.8720     & 0.9400   
  & 0.8090     & 0.8920   
  & 0.7448     & 0.8334  \\
Intern2.5-VL-78B~\cite{chen2024expanding}        
  & 0.8368     & 0.8795   
  & 0.7123     & 0.7652   
  & 0.3379     & 0.3735  \\ 
\midrule
\multicolumn{7}{l}{\textit{Expert VLMs}}                                                                                                              \\
Nougat~\cite{blecher2023nougat}                  
  & 0.3856     & 0.4308   
  & 0.1193     & 0.1357   
  & 0.0424     & 0.0527  \\ 
\midrule
\multicolumn{7}{l}{\textit{Our Results}}                                                                                                              \\

Intern2-VL-1B-VSGRPO         
  & 0.8959     & 0.9358   
  & 0.8604     & 0.8988   
  & 0.8054     & 0.8625  \\

Qwen2.5-VL-3B-VSGRPO         
  & \textbf{0.8997}     & \textbf{0.9405}   
  & \textbf{0.9004}     & \textbf{0.9427}   
  & \textbf{0.8673}     & \textbf{0.9218}  \\
\bottomrule
\end{tabular}
}
  \vspace{-5pt}
\end{table}

To evaluate the generalization capability of our method, we conduct additional experiments on an external benchmark dataset introduced in~\cite{Jiang2024LATTEIL}. The results are presented in \Cref{tab:3}. Manual inspection reveals that this dataset is primarily composed of simple tables with limited structural complexity. Consequently, the performance trends largely mirror those observed in the simple-table subsets reported in \Cref{tab:1} and \Cref{tab:2}.

Once again, our method, Qwen2.5-VL-3B-VSGRPO, achieves superior performance, outperforming both task-specific baselines for table image to LaTeX generation\footnote{The LATTE model proposed in~\cite{Jiang2024LATTEIL} is not publicly available. We compute results based on the authors’ released outputs and apply our own metric calculations.} and general-purpose multimodal large language models. These results underscore the model’s strong generalization capability.


\begin{table}[htbp]
  \vspace{-5pt}
  \centering
  \caption{Experimental comparison on external dataset~\cite{Jiang2024LATTEIL} on CW-SSIM and TEDS-Structure.}
  \label{tab:3}
  \resizebox{0.5\textwidth}{!}{
  \begin{tabular}{lcc}
    \toprule
    Models & CW-SSIM & TEDS-Structure \\
    \midrule
    LATTE~\cite{Jiang2024LATTEIL} & 0.7615 & 0.9445 \\
    GPT4o~\cite{openai2025gpt4o} & 0.6897 & 0.8568 \\
    Qwen2.5-VL-72B~\cite{bai2025qwen2} & 0.7176 & 0.8915 \\
    Intern2.5-VL-78B~\cite{chen2024expanding} & 0.7696 & 0.9009 \\
    Qwen2.5-VL-3B-VSGRPO & \textbf{0.8225} & \textbf{0.9461} \\
    \bottomrule
  \end{tabular}
  }
    \vspace{-5pt}
\end{table}

\subsection{Human Evaluation}

\label{human_evaluation}
To complement automated metrics and better capture perceived visual quality, we conduct a human preference study on 200 randomly selected tables (50 simple, 50 medium, 100 complex), as shown in \Cref{human}. For each case, rendered outputs from four models are displayed anonymously alongside the ground-truth image. Multiple human assessors independently vote on the most visually similar result, and the final decision is determined by majority voting. As shown in \Cref{tab:human_evaluation}, Qwen2.5-VL-3B-VSGRPO receives the highest number of votes across all difficulty levels, clearly outperforming other models in terms of visual and structural fidelity.

\begin{table}[ht]
  \vspace{-5pt}
\centering
\caption{Results of human evaluation.}
  \resizebox{0.52\textwidth}{!}{
\begin{tabular}{lccc}
\toprule
Models & Simple & Medium & Complex \\
\midrule
GPT4o & 5 & 2 & 2 \\
Mathpix & 19 & 2 & 10 \\
Qwen2.5-VL-3B-SFT & 29 & 28 & 56 \\
Qwen2.5-VL-3B-VSGRPO & \textbf{42} & \textbf{37} & \textbf{70} \\
\bottomrule
\end{tabular}
}
\label{tab:human_evaluation}
  \vspace{-5pt}
\end{table}

\subsection{Ablation Study}

\label{ablation_study}
To validate the effectiveness and robustness of our proposed method, we conduct a series of ablation studies focusing on three aspects: the impact of data selection strategies, the contribution of individual reward components, and the necessity of staged training. All experiments are evaluated on the complex table subset.

\textbf{Evaluation on the Dataset Selection Strategy for VSGRPO.}\quad   As shown in Table~\ref{tab:dataset_selection}, we compare different strategies for constructing the reinforcement learning (RL) training set. Specifically, we evaluate three variants of Qwen2.5-VL-3B fine-tuned with VSGRPO: (1) using only simple tables (\texttt{-Simple}), (2) using a balanced mixture of simple, medium, and complex tables (\texttt{-Mixed-Data}), and (3) using only complex tables (\texttt{-VSGRPO}). The results clearly demonstrate that restricting the RL fine-tuning data to complex tables leads to the best overall performance across all metrics. This validates our design choice of focusing on structurally difficult examples during reinforcement learning to better generalize across complexity levels.

\begin{table}[htbp]
  \vspace{-5pt}
  \centering
  \caption{Ablation experiments on the dataset selection for VSGRPO.}
  \label{tab:dataset_selection}
  \resizebox{0.8\textwidth}{!}{
  \begin{tabular}{lcccc}
    \toprule
    Models & CW-SSIM & \makecell{Compile ratio} & \makecell{TEDS} & \makecell{TEDS-Structure} \\
    \midrule
    Qwen2.5-VL-3B-VSGRPO-Simple       & 0.5993 & 0.9861 & 0.8614 & 0.9113 \\
    Qwen2.5-VL-3B-VSGRPO-Mixed-Data & 0.6107 & 0.9861 & 0.8614 & 0.9136 \\
    Qwen2.5-VL-3B-VSGRPO            & \textbf{0.6145} & \textbf{0.9917} & \textbf{0.8673} & \textbf{0.9218} \\
    \bottomrule
  \end{tabular}
  }
  \vspace{-5pt}
\end{table}

\textbf{Evaluation on the Reward Design in VSGRPO.}\quad Table~\ref{tab:reward_ablation} presents the effectiveness of the two reward components used in our RL framework—TEDS-Structure and CW-SSIM. Adding either reward individually to the base model leads to noticeable performance gains over the SFT-only baseline, demonstrating that both structure-level accuracy and visual similarity are important for improving LaTeX generation quality. The best performance is achieved when both reward signals are combined, suggesting they are complementary in guiding the model toward faithful and well-aligned outputs.

\begin{table}[htbp]
  \vspace{-5pt}
  \centering
  \caption{Ablation experiments on the reward design.}
  \label{tab:reward_ablation}
  \resizebox{0.8\textwidth}{!}{
  \begin{tabular}{lcccc}
    \toprule
    Models & CW-SSIM & \makecell{Compile ratio} & \makecell{TEDS} & \makecell{TEDS-Structure} \\
    \midrule
    Qwen2.5-VL-3B-SFT                      & 0.5806 & 0.9889 & 0.8481 & 0.9047 \\
    Qwen2.5-VL-3B-GRPO-TEDS-Structure       & 0.5925 & 0.9889 & 0.8608 & 0.9155 \\
    Qwen2.5-VL-3B-GRPO-CW-SSIM             & 0.6064 & 0.9889 & 0.8607 & 0.9133 \\
    Qwen2.5-VL-3B-VSGRPO                     & \textbf{0.6145} & \textbf{0.9917} & \textbf{0.8673} & \textbf{0.9218} \\
    \bottomrule
  \end{tabular}
  }
    \vspace{-5pt}
\end{table}

\textbf{Evaluation on the Necessicity of SFT.}\quad 
To verify the necessity of SFT before reinforcement fine-tuning, we perform one epoch of reinforcement learning directly on the pre-trained Qwen2.5-VL-3B model (VSGRPO without SFT). As shown in Table~\ref{tab:w/o_sft}, the performance of the model without SFT initialization is significantly lower across all metrics, indicating that SFT is essential to provide a reasonable starting point for subsequent RL-based optimization.

\begin{table}[htbp]
  \vspace{-5pt}
  \centering
  \caption{Ablation experiments on the effectiveness of SFT.}
  \label{tab:w/o_sft}
  \resizebox{0.8\textwidth}{!}{
  \begin{tabular}{lcccc}
    \toprule
    Models & CW-SSIM & \makecell{Compile ratio} & TEDS & \makecell{TEDS-Structure} \\
    \midrule
    Qwen2.5-VL-3B-VSGRPO w/o SFT & 0.4695 & 0.9668 & 0.6884 & 0.8167 \\
    Qwen2.5-VL-3B-VSGRPO & \textbf{0.6145} & \textbf{0.9917} & \textbf{0.8673} & \textbf{0.9218} \\
    \bottomrule
  \end{tabular}
  }
    \vspace{-5pt}
\end{table}

\section{Conclusion and Limitations}

 Our work tackled the challenge of converting table images into syntactically correct, publication-quality LaTeX code by integrating vision-language modeling with targeted reinforcement learning. We leveraged a pre-trained multimodal large language model (MLLM), fine-tuned it on a diverse corpus of scientific table images, and further enhanced it through a dual-reward scheme: one reward evaluated structural integrity using TEDS-Structure, while the other measured visual fidelity via a refined CW-SSIM on rendered outputs. By jointly optimizing these objectives, the model was able to accurately capture complex table layouts—including nested headers, merged cells, and mathematical expressions—and produce outputs that closely matched the original visual appearance.\

\textbf{Limitations.}\quad Although VSGRPO effectively improved MLLM performance on complex tables, it introduced notable computational overhead during training. Specifically, each LaTeX output had to be rendered into a PDF and then converted to a PNG image for CW-SSIM computation—a time-consuming process that created a training bottleneck, even with multi-threading. Due to this overhead and limited GPU resources, we trained VSGRPO on only 5,936 complex tables.

\bibliographystyle{unsrtnat}
\bibliography{reference}

\begin{thebibliography}{38}
\providecommand{\natexlab}[1]{#1}
\providecommand{\url}[1]{\texttt{#1}}
\expandafter\ifx\csname urlstyle\endcsname\relax
  \providecommand{\doi}[1]{doi: #1}\else
  \providecommand{\doi}{doi: \begingroup \urlstyle{rm}\Url}\fi

\bibitem[Zhong et~al.(2020)Zhong, ShafieiBavani, and Jimeno~Yepes]{zhong2020image}
Xu~Zhong, Elaheh ShafieiBavani, and Antonio Jimeno~Yepes.
\newblock Image-based table recognition: data, model, and evaluation.
\newblock In \emph{ECCV}. Springer, 2020.

\bibitem[Ye et~al.(2021)Ye, Qi, He, Chen, Gu, Gao, and Xiao]{ye2021pingan}
Jiaquan Ye, Xianbiao Qi, Yelin He, Yihao Chen, Dengyi Gu, Peng Gao, and Rong Xiao.
\newblock Pingan-vcgroup's solution for icdar 2021 competition on scientific literature parsing task b: table recognition to html.
\newblock \emph{arXiv}, 2021.

\bibitem[Nassar et~al.(2022)Nassar, Livathinos, Lysak, and Staar]{nassar2022tableformer}
Ahmed Nassar, Nikolaos Livathinos, Maksym Lysak, and Peter Staar.
\newblock Tableformer: Table structure understanding with transformers.
\newblock In \emph{CVPR}, 2022.

\bibitem[Huang et~al.(2023{\natexlab{a}})Huang, Lu, Chen, Li, Xie, Zhu, Gao, and Peng]{huang2023improving}
Yongshuai Huang, Ning Lu, Dapeng Chen, Yibo Li, Zecheng Xie, Shenggao Zhu, Liangcai Gao, and Wei Peng.
\newblock Improving table structure recognition with visual-alignment sequential coordinate modeling.
\newblock In \emph{CVPR}, 2023{\natexlab{a}}.

\bibitem[Shen et~al.(2023)Shen, Gao, Wei, Qiao, Zhou, Li, and Cheng]{shen2023divide}
Huawen Shen, Xiang Gao, Jin Wei, Liang Qiao, Yu~Zhou, Qiang Li, and Zhanzhan Cheng.
\newblock Divide rows and conquer cells: Towards structure recognition for large tables.
\newblock In \emph{IJCAI}, 2023.

\bibitem[Jiang et~al.(2024)Jiang, Liang, Wang, Wang, and Tan]{Jiang2024LATTEIL}
Nan Jiang, Shanchao Liang, Chengxiao Wang, Jiannan Wang, and Lin Tan.
\newblock Latte: Improving latex recognition for tables and formulae with iterative refinement.
\newblock In \emph{AAAI}, 2024.

\bibitem[Xia et~al.(2025)Xia, Zhou, Feng, Liu, Chen, Zhang, and Yan]{xia2025latexnet}
Renqiu Xia, Hongbin Zhou, Ziming Feng, Huanxi Liu, Boan Chen, Bo~Zhang, and Junchi Yan.
\newblock Latexnet: A specialized model for converting visual tables and equations to latex code.
\newblock In \emph{ICASSP 2025-2025 IEEE International Conference on Acoustics, Speech and Signal Processing (ICASSP)}, 2025.

\bibitem[Shao et~al.(2024{\natexlab{a}})Shao, Wang, Zhu, Xu, Song, Zhang, Li, Wu, and Guo]{Shao2024DeepSeekMathPT}
Zhihong Shao, Peiyi Wang, Qihao Zhu, Runxin Xu, Jun-Mei Song, Mingchuan Zhang, Y.~K. Li, Yu~Wu, and Daya Guo.
\newblock Deepseekmath: Pushing the limits of mathematical reasoning in open language models.
\newblock \emph{ArXiv}, 2024{\natexlab{a}}.

\bibitem[Zhong et~al.(2019)Zhong, Shafieibavani, and Jimeno-Yepes]{Zhong2019ImagebasedTR}
Xu~Zhong, Elaheh Shafieibavani, and Antonio Jimeno-Yepes.
\newblock Image-based table recognition: data, model, and evaluation.
\newblock In \emph{ECCV}, 2019.

\bibitem[Huang et~al.(2023{\natexlab{b}})Huang, Lu, Chen, Li, Xie, Zhu, Gao, and Peng]{Huang2023ImprovingTS}
Yongshuai Huang, Ning Lu, Dapeng Chen, Yibo Li, Zecheng Xie, Shenggao Zhu, Liangcai Gao, and Wei Peng.
\newblock Improving table structure recognition with visual-alignment sequential coordinate modeling.
\newblock \emph{CVPR}, 2023{\natexlab{b}}.

\bibitem[Schreiber et~al.(2017)Schreiber, Agne, Wolf, Dengel, and Ahmed]{schreiber2017deepdesrt}
Sebastian Schreiber, Stefan Agne, Ivo Wolf, Andreas Dengel, and Sheraz Ahmed.
\newblock Deepdesrt: Deep learning for detection and structure recognition of tables in document images.
\newblock In \emph{ICDAR}, 2017.

\bibitem[Paliwal et~al.(2019)Paliwal, Vishwanath, Rahul, Sharma, and Vig]{paliwal2019tablenet}
Shubham~Singh Paliwal, D~Vishwanath, Rohit Rahul, Monika Sharma, and Lovekesh Vig.
\newblock Tablenet: Deep learning model for end-to-end table detection and tabular data extraction from scanned document images.
\newblock In \emph{ICDAR}, 2019.

\bibitem[Tensmeyer et~al.(2019)Tensmeyer, Morariu, Price, Cohen, and Martinez]{tensmeyer2019deep}
Chris Tensmeyer, Vlad~I Morariu, Brian Price, Scott Cohen, and Tony Martinez.
\newblock Deep splitting and merging for table structure decomposition.
\newblock In \emph{ICDAR}, 2019.

\bibitem[Guo et~al.(2022)Guo, Yu, Lv, Zhang, Li, Wang, Yao, Liu, and Wang]{guo2022trust}
Zengyuan Guo, Yuechen Yu, Pengyuan Lv, Chengquan Zhang, Haojie Li, Zhihui Wang, Kun Yao, Jingtuo Liu, and Jingdong Wang.
\newblock Trust: An accurate and end-to-end table structure recognizer using splitting-based transformers.
\newblock \emph{arXiv}, 2022.

\bibitem[Zhang et~al.(2022)Zhang, Zhang, Du, and Wang]{zhang2022split}
Zhenrong Zhang, Jianshu Zhang, Jun Du, and Fengren Wang.
\newblock Split, embed and merge: An accurate table structure recognizer.
\newblock \emph{PR}, 2022.

\bibitem[Ma et~al.(2023)Ma, Lin, Sun, and Huo]{ma2023robust}
Chixiang Ma, Weihong Lin, Lei Sun, and Qiang Huo.
\newblock Robust table detection and structure recognition from heterogeneous document images.
\newblock \emph{PR}, 2023.

\bibitem[Lyu et~al.(2023)Lyu, Ma, Wang, Yu, Zhang, Yao, Xue, and Wang]{lyu2023gridformer}
Pengyuan Lyu, Weihong Ma, Hongyi Wang, Yuechen Yu, Chengquan Zhang, Kun Yao, Yang Xue, and Jingdong Wang.
\newblock Gridformer: Towards accurate table structure recognition via grid prediction.
\newblock In \emph{ACM MM}, 2023.

\bibitem[Raja et~al.(2020)Raja, Mondal, and Jawahar]{raja2020table}
Sachin Raja, Ajoy Mondal, and CV~Jawahar.
\newblock Table structure recognition using top-down and bottom-up cues.
\newblock In \emph{ECCV}, 2020.

\bibitem[Liu et~al.(2021)Liu, Li, Liu, Jiang, Liu, Ren, and Ji]{liu2021show}
Hao Liu, Xin Li, Bing Liu, Deqiang Jiang, Yinsong Liu, Bo~Ren, and Rongrong Ji.
\newblock Show, read and reason: Table structure recognition with flexible context aggregator.
\newblock In \emph{ACM MM}, 2021.

\bibitem[Wan et~al.(2024)Wan, Song, Yu, Liu, Cheng, Huang, Bai, Yao, and Yang]{wan2024omniparser}
Jianqiang Wan, Sibo Song, Wenwen Yu, Yuliang Liu, Wenqing Cheng, Fei Huang, Xiang Bai, Cong Yao, and Zhibo Yang.
\newblock Omniparser: A unified framework for text spotting key information extraction and table recognition.
\newblock In \emph{CVPR}, 2024.

\bibitem[Chi et~al.(2019)Chi, Huang, Xu, Yu, Yin, and Mao]{chi2019complicated}
Zewen Chi, Heyan Huang, Heng-Da Xu, Houjin Yu, Wanxuan Yin, and Xian-Ling Mao.
\newblock Complicated table structure recognition.
\newblock \emph{arXiv}, 2019.

\bibitem[Blecher et~al.(2023)Blecher, Cucurull, Scialom, and Stojnic]{blecher2023nougat}
Lukas Blecher, Guillem Cucurull, Thomas Scialom, and Robert Stojnic.
\newblock Nougat: Neural optical understanding for academic documents.
\newblock \emph{arXiv}, 2023.

\bibitem[Sun et~al.(2023)Sun, Shen, Cao, Liu, Li, Shen, Gan, Gui, Wang, Yang, Keutzer, and Darrell]{Sun2023AligningLM}
Zhiqing Sun, Sheng Shen, Shengcao Cao, Haotian Liu, Chunyuan Li, Yikang Shen, Chuang Gan, Liangyan Gui, Yu-Xiong Wang, Yiming Yang, Kurt Keutzer, and Trevor Darrell.
\newblock Aligning large multimodal models with factually augmented rlhf.
\newblock \emph{ArXiv}, 2023.

\bibitem[Yu et~al.(2023)Yu, Yao, Zhang, He, Han, Cui, Hu, Liu, Zheng, Sun, and Chua]{Yu2023RLHFVTT}
Tianyu Yu, Yuan Yao, Haoye Zhang, Taiwen He, Yifeng Han, Ganqu Cui, Jinyi Hu, Zhiyuan Liu, Hai-Tao Zheng, Maosong Sun, and Tat-Seng Chua.
\newblock Rlhf-v: Towards trustworthy mllms via behavior alignment from fine-grained correctional human feedback.
\newblock \emph{CVPR}, 2023.

\bibitem[Huang et~al.(2025)Huang, Jia, Zhai, Cao, Ye, Zhao, Xu, Hu, and Lin]{Huang2025VisionR1IR}
Wenxuan Huang, Bohan Jia, Zijie Zhai, Shaoshen Cao, Zheyu Ye, Fei Zhao, Zhe Xu, Yao Hu, and Shaohui Lin.
\newblock Vision-r1: Incentivizing reasoning capability in multimodal large language models.
\newblock \emph{ArXiv}, 2025.

\bibitem[Shen et~al.(2025{\natexlab{a}})Shen, Liu, Li, Fang, Ma, Liao, Shen, Zhang, Zhao, Zhang, Xu, and Zhao]{Shen2025VLMR1AS}
Haozhan Shen, Peng Liu, Jingcheng Li, Chunxin Fang, Yibo Ma, Jiajia Liao, Qiaoli Shen, Zilun Zhang, Kangjia Zhao, Qianqian Zhang, Ruochen Xu, and Tiancheng Zhao.
\newblock Vlm-r1: A stable and generalizable r1-style large vision-language model.
\newblock \emph{ArXiv}, 2025{\natexlab{a}}.

\bibitem[Papineni et~al.(2002)Papineni, Roukos, Ward, and Zhu]{Papineni2002BleuAM}
Kishore Papineni, Salim Roukos, Todd Ward, and Wei-Jing Zhu.
\newblock Bleu: a method for automatic evaluation of machine translation.
\newblock In \emph{Annual Meeting of the Association for Computational Linguistics}, 2002.

\bibitem[Gao et~al.(2011)Gao, Rehman, and Wang]{CWSSIM}
Yang Gao, Abdul Rehman, and Zhou Wang.
\newblock Cw-ssim based image classification.
\newblock In \emph{2011 18th IEEE International Conference on Image Processing}, 2011.

\bibitem[Schmidt(1907)]{schmidt1907theorie}
Erhard Schmidt.
\newblock Zur theorie der linearen und nichtlinearen integralgleichungen.
\newblock \emph{Mathematische Annalen}, 1907.

\bibitem[Shao et~al.(2024{\natexlab{b}})Shao, Wang, Zhu, Xu, Song, Bi, Zhang, Zhang, Li, Wu, et~al.]{shao2024deepseekmath}
Zhihong Shao, Peiyi Wang, Qihao Zhu, Runxin Xu, Junxiao Song, Xiao Bi, Haowei Zhang, Mingchuan Zhang, YK~Li, Y~Wu, et~al.
\newblock Deepseekmath: Pushing the limits of mathematical reasoning in open language models.
\newblock \emph{arXiv}, 2024{\natexlab{b}}.

\bibitem[Schulman et~al.(2017)Schulman, Wolski, Dhariwal, Radford, and Klimov]{schulman2017proximal}
John Schulman, Filip Wolski, Prafulla Dhariwal, Alec Radford, and Oleg Klimov.
\newblock Proximal policy optimization algorithms.
\newblock \emph{arXiv preprint arXiv:1707.06347}, 2017.

\bibitem[Zheng et~al.(2024)Zheng, Zhang, Zhang, Ye, Luo, Feng, and Ma]{zheng2024llamafactory}
Yaowei Zheng, Richong Zhang, Junhao Zhang, Yanhan Ye, Zheyan Luo, Zhangchi Feng, and Yongqiang Ma.
\newblock Llamafactory: Unified efficient fine-tuning of 100+ language models.
\newblock In \emph{Proceedings of the 62nd Annual Meeting of the Association for Computational Linguistics (Volume 3: System Demonstrations)}, 2024.

\bibitem[Shen et~al.(2025{\natexlab{b}})Shen, Liu, Li, Fang, Ma, Liao, Shen, Zhang, Zhao, Zhang, Xu, and Zhao]{shen2025vlm}
Haozhan Shen, Peng Liu, Jingcheng Li, Chunxin Fang, Yibo Ma, Jiajia Liao, Qiaoli Shen, Zilun Zhang, Kangjia Zhao, Qianqian Zhang, Ruochen Xu, and Tiancheng Zhao.
\newblock Vlm-r1: A stable and generalizable r1-style large vision-language model.
\newblock \emph{arXiv preprint arXiv:2504.07615}, 2025{\natexlab{b}}.

\bibitem[Zhao et~al.(2024)Zhao, Huang, Hu, Wang, Mao, Zhang, Jiang, Wu, Ai, Wang, Zhou, and Chen]{zhao2024swiftascalablelightweightinfrastructure}
Yuze Zhao, Jintao Huang, Jinghan Hu, Xingjun Wang, Yunlin Mao, Daoze Zhang, Zeyinzi Jiang, Zhikai Wu, Baole Ai, Ang Wang, Wenmeng Zhou, and Yingda Chen.
\newblock Swift:a scalable lightweight infrastructure for fine-tuning, 2024.

\bibitem[{Mathpix, Inc.}(2025)]{mathpix2025}
{Mathpix, Inc.}
\newblock Mathpix — ai‑powered text recognition.
\newblock \url{https://mathpix.com/}, 2025.

\bibitem[{OpenAI}(2025)]{openai2025gpt4o}
{OpenAI}.
\newblock Gpt‑4o: Gpt‑4 with vision capabilities.
\newblock \url{https://openai.com/research/gpt-4o}, 2025.

\bibitem[Bai et~al.(2025)Bai, Chen, Liu, Wang, Ge, Song, Dang, Wang, Wang, Tang, et~al.]{bai2025qwen2}
Shuai Bai, Keqin Chen, Xuejing Liu, Jialin Wang, Wenbin Ge, Sibo Song, Kai Dang, Peng Wang, Shijie Wang, Jun Tang, et~al.
\newblock Qwen2. 5-vl technical report.
\newblock \emph{arXiv preprint arXiv:2502.13923}, 2025.

\bibitem[Chen et~al.(2024)Chen, Wang, Cao, Liu, Gao, Cui, Zhu, Ye, Tian, Liu, et~al.]{chen2024expanding}
Zhe Chen, Weiyun Wang, Yue Cao, Yangzhou Liu, Zhangwei Gao, Erfei Cui, Jinguo Zhu, Shenglong Ye, Hao Tian, Zhaoyang Liu, et~al.
\newblock Expanding performance boundaries of open-source multimodal models with model, data, and test-time scaling.
\newblock \emph{arXiv preprint arXiv:2412.05271}, 2024.

\end{thebibliography}
\newpage
\appendix

\section{Case Study}
\label{case}
In this section, we demonstrate the limitations of relevant metrics through result visualizations, and the capability of our method VSGRPO to generate such complex table and analyze the effectiveness of our approach.

\begin{figure}[ht]
    \centering
    \includegraphics[width=1\linewidth]{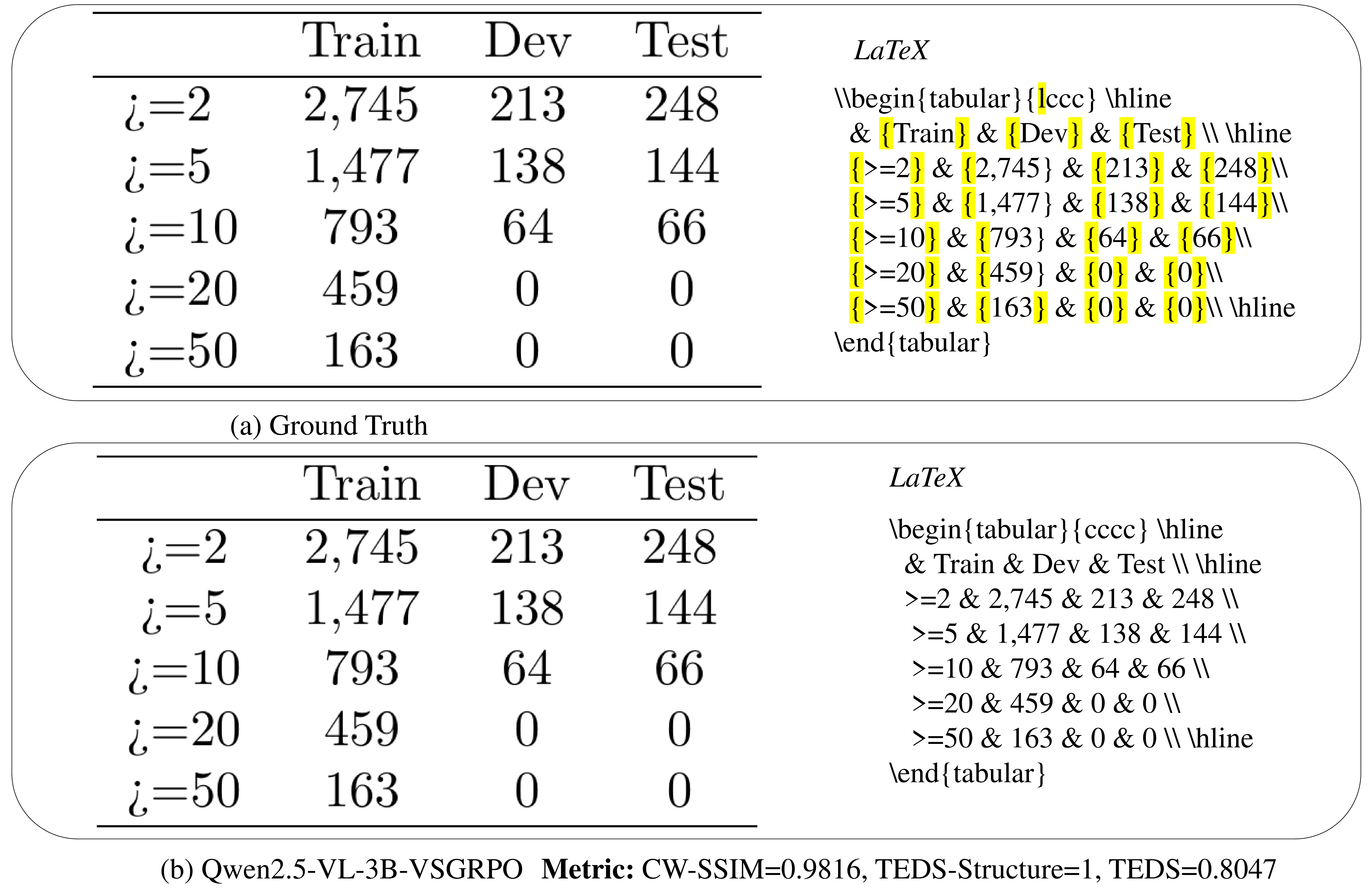}
    \caption{Example 1: LaTeX code ambiguity.}
    \label{fig:case3}
\end{figure}

\begin{figure}[ht]
    \centering
    \includegraphics[width=1\linewidth]{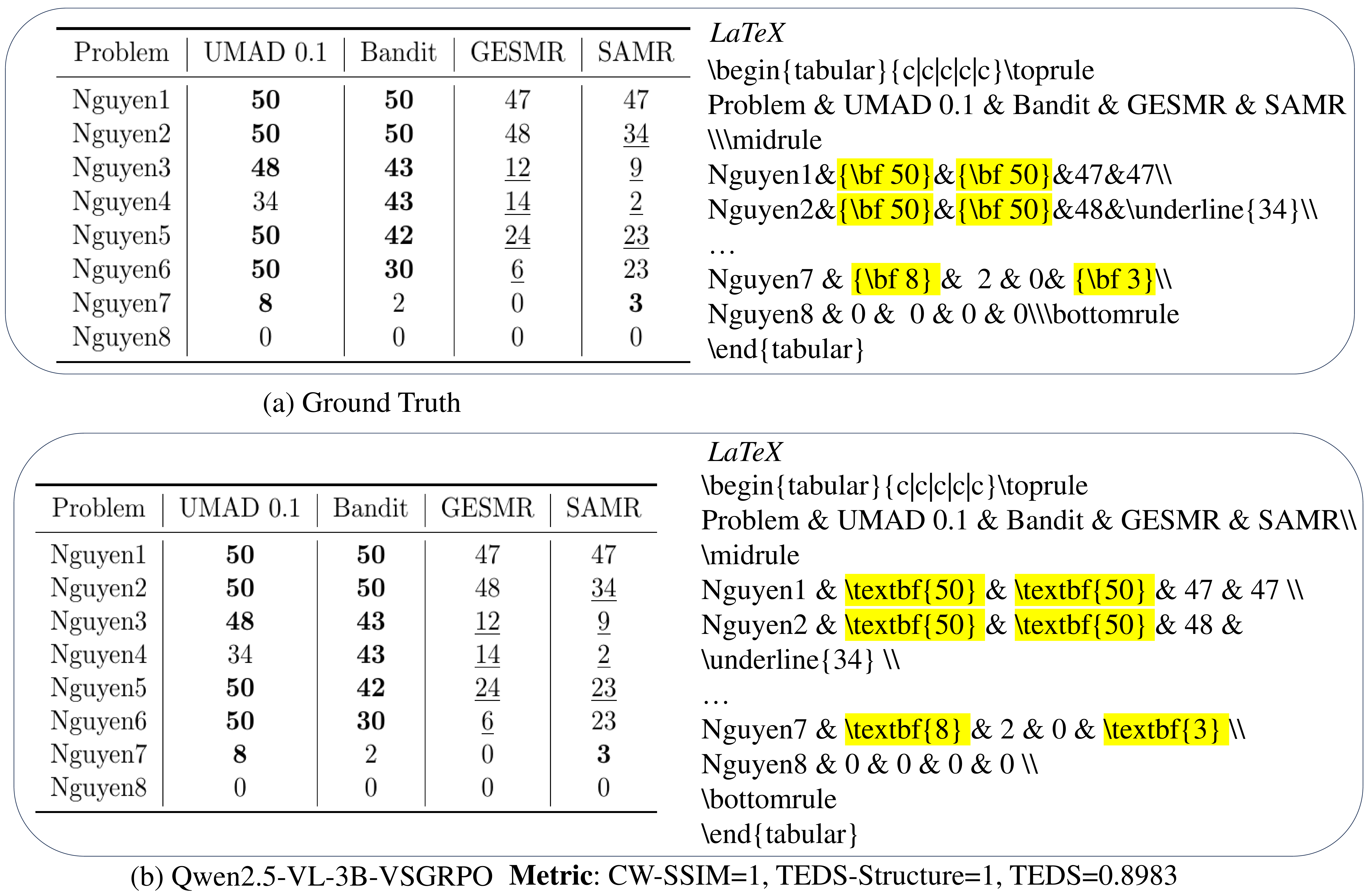}
    \caption{Example 2: LaTeX code ambiguity.}
    \label{fig:case2}
\end{figure}

\textbf{The Limitation of Metric.}\quad We illustrate LaTeX‑level ambiguity with two visually identical table renderings whose TEDS scores nonetheless differ. In \Cref{fig:case3} (TEDS 0.8047), the ground‑truth code wraps every cell’s contents in an empty group \{\,\}, whereas the model output omits these no‑op braces. Although neither variation alters the final rendering, they change the underlying token sequence and thus reduce the TEDS score. In both figures, the relevant LaTeX code differences are highlighted in yellow. In \Cref{fig:case2} (TEDS 0.8983), the sole divergence lies in the use of different bold commands. 

\begin{figure}[ht]
    \centering
    \includegraphics[width=1\linewidth]{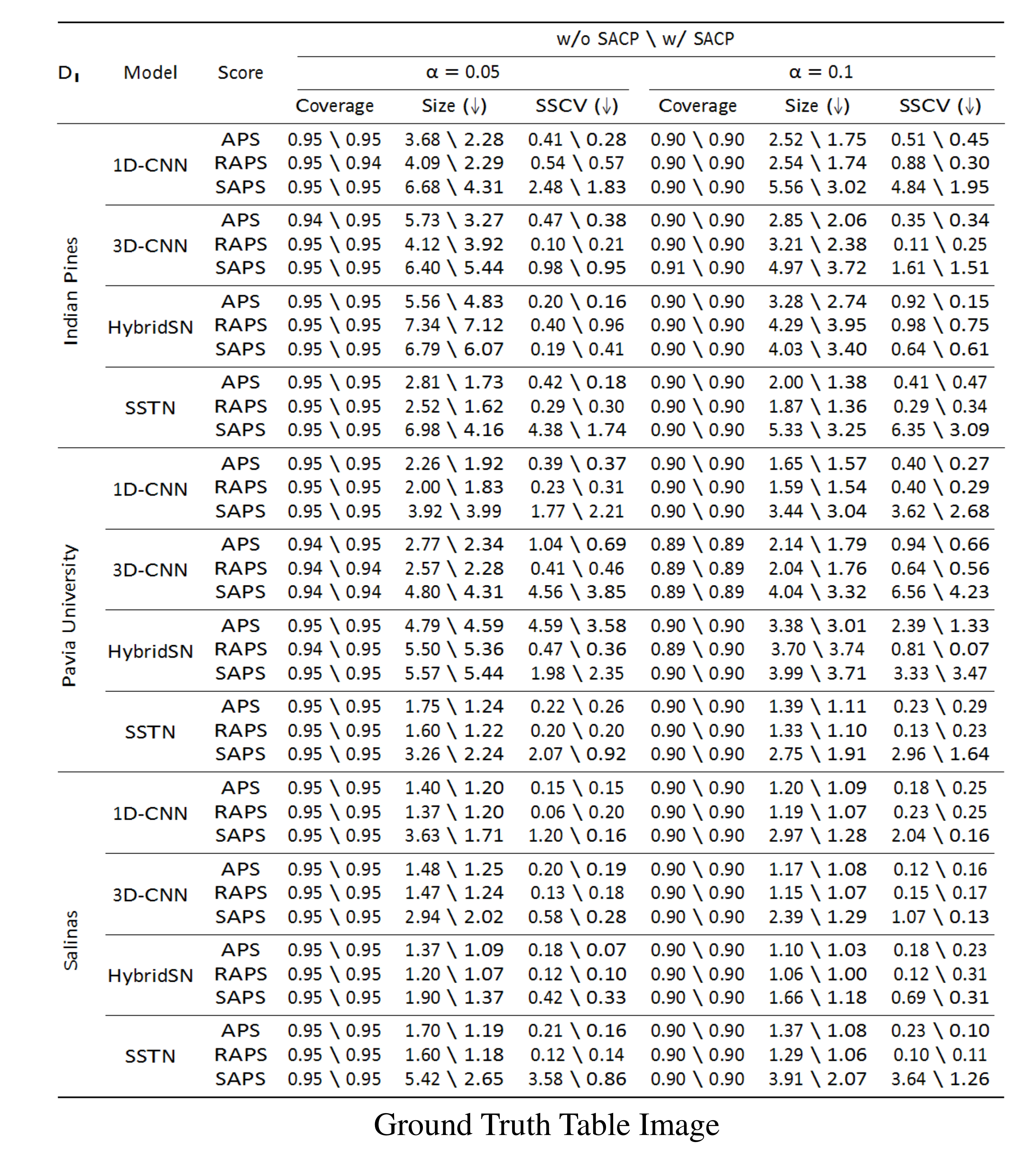}
    \caption{Ground truth example of complex table.}
    \label{fig:complex_gt}
\end{figure}

\begin{figure}[ht]
    \centering
    \includegraphics[width=1\linewidth]{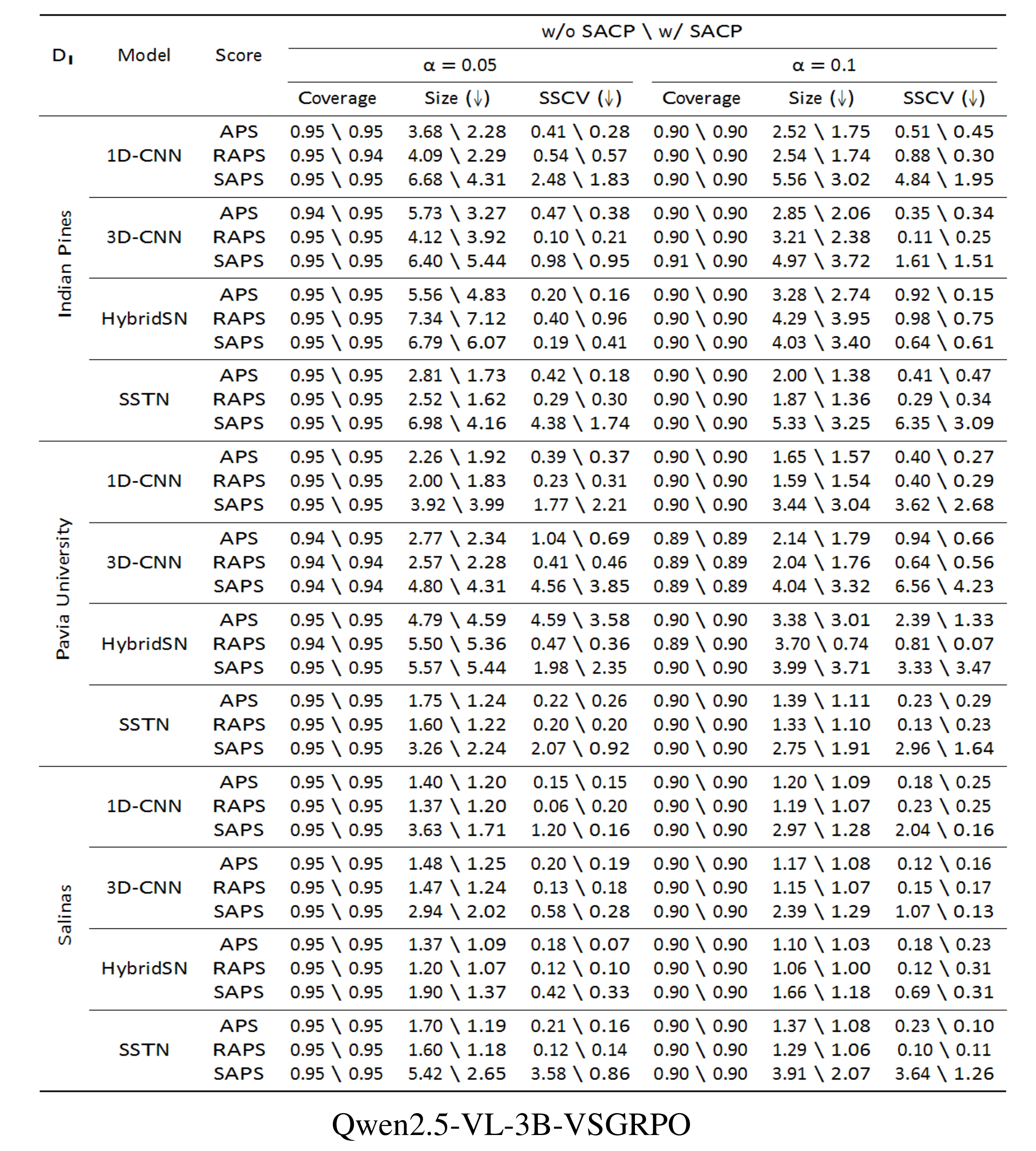}
    \caption{Prediction example of complex table.}
    \label{fig:complex_pre}
\end{figure}

\textbf{Visualisation of Complex Tables.}\quad As shown in \Cref{fig:complex_gt} and \Cref{fig:complex_pre}, they are the complex table image from the ground truth and the table image rendered from the LaTeX generated by our method VSGRPO, respectively.

\begin{figure}[ht]
    \centering
    \includegraphics[width=0.95\linewidth]{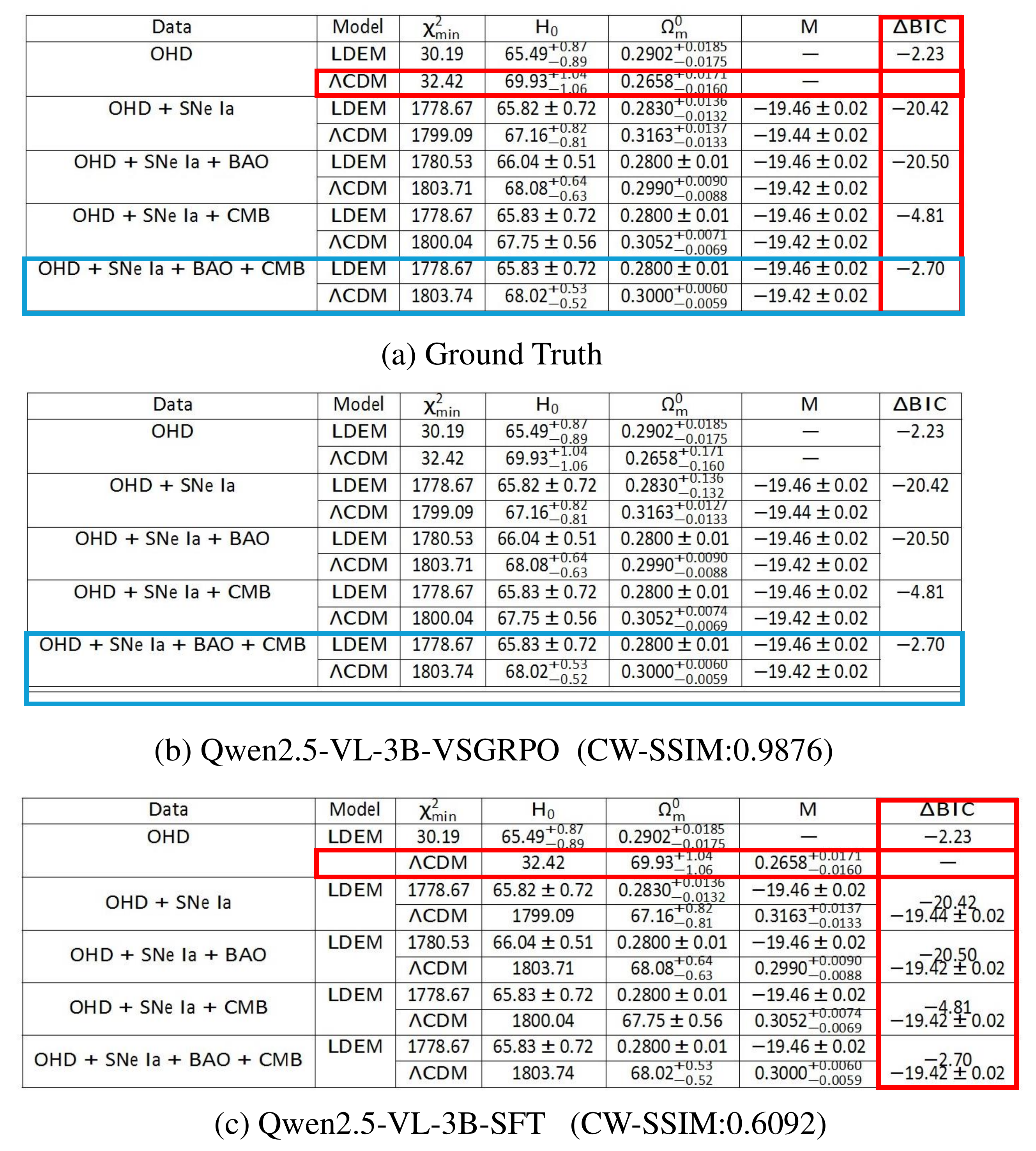}
    \caption{Visualization of result comparisons. (a) Ground Truth refers to the ground truth table image from the simple testing dataset; (b) Qwen2.5-VL-3B-VSGRPO represents the table image rendered from LaTeX generated by the Qwen2.5-VL-3B model trained with our VSGRPO method; (c) Qwen2.5-VL-3B-SFT represents the table image rendered from LaTeX generated by the Qwen2.5-VL-3B model trained with SFT. The corresponding CW-SSIM scores are reported. Blue boxes highlight examples where Qwen2.5-VL-3B-VSGRPO differs from the ground truth, and red boxes highlight examples where Qwen2.5-VL-3B-SFT differs from the ground truth.}
    \label{fig:case1}
\end{figure}

\textbf{Visualisation of effectiveness.}\quad As shown in \Cref{fig:case1}, our method VSGRPO improves the quality of the LaTeX generated by SFT, and the CW-SSIM score also reflects the visual similarity between the images.

\section{Metric}
\label{metric}
\textbf{CW-SSIM.}\quad The specific formula is as follows:
\begin{equation}
\mathrm{CW\text{-}SSIM}(X,Y)
=\frac{1}{4}\sum_{i\in\{A,H,V,D\}}
\mathrm{SSIM}\bigl(c_X^i,\,c_Y^i\bigr),
\end{equation}
where for each sub-band \(i\):
\begin{equation}
\mathrm{SSIM}(c_X^i,c_Y^i)
=
\displaystyle
\frac{\bigl(2\mu_{X_i}\mu_{Y_i}+C_1\bigr)\,\bigl(2\sigma_{X_iY_i}+C_2\bigr)}
{\bigl(\mu_{X_i}^2+\mu_{Y_i}^2+C_1\bigr)\,\bigl(\sigma_{X_i}^2+\sigma_{Y_i}^2+C_2\bigr)}.
\end{equation}

Let \(X\) and \(Y\) be two aligned grayscale table images trimmed to even dimensions.  Apply a one-level Haar wavelet to each, yielding four sub-bands \(c_X^i\) and \(c_Y^i\) for \(i\!=\!A\) (approximation), \(H\) (horizontal detail), \(V\) (vertical detail), and \(D\) (diagonal detail).  For each sub-band \(i\), let \(\mu_{X_i}\) and \(\mu_{Y_i}\) be the pixel-wise means, \(\sigma_{X_i}^2\) and \(\sigma_{Y_i}^2\) the variances, and \(\sigma_{X_iY_i}\) the covariance.  Constants \(C_1=(K_1L)^2\) and \(C_2=(K_2L)^2\) stabilize the denominator (\(L\) is 255.0, \(K_1\) is 0.01 and \(K_2\) is 0.03). 

\textbf{TEDS-Structure.}\quad The specific formula is as follows:
\begin{equation}\label{eq6}
\mathrm{TEDS\text{-}Structure}
= 1 - \frac{\mathrm{TED}_{\mathrm{structure}}}
           {\max\bigl(|T_{\mathrm{pred}}|,\;|T_{\mathrm{gt}}|\bigr)},
\end{equation}
$\lvert T_{\mathrm{pred}}\rvert$ denotes the total number of nodes in the structural tree parsed from the predicted table, and $\lvert T_{\mathrm{gt}}\rvert$ denotes the total number of nodes in the structural tree parsed from the ground-truth table.

\textbf{TEDS.}\quad The Tree Edit Distance–based Similarity (TEDS) computation extends TEDS‑Structure by first calculating the total edit distance: $\mathrm{TED} = \mathrm{TED}_{\mathrm{structure}} + \mathrm{TED}_{\mathrm{content}}$, and finally normalizing by the larger of the two tree sizes to yield the similarity score. The specific formula is as follows:
\begin{equation}\label{eq8}
\mathrm{TEDS}
= 1 - \frac{\mathrm{TED}}{\max\bigl(\lvert T_{\mathrm{pred}}\rvert,\;\lvert T_{\mathrm{gt}}\rvert\bigr)}.
\end{equation}

\section{Human Evaluation}
\label{human}
\begin{figure}[ht]
    \centering
    \includegraphics[width=0.95\linewidth]{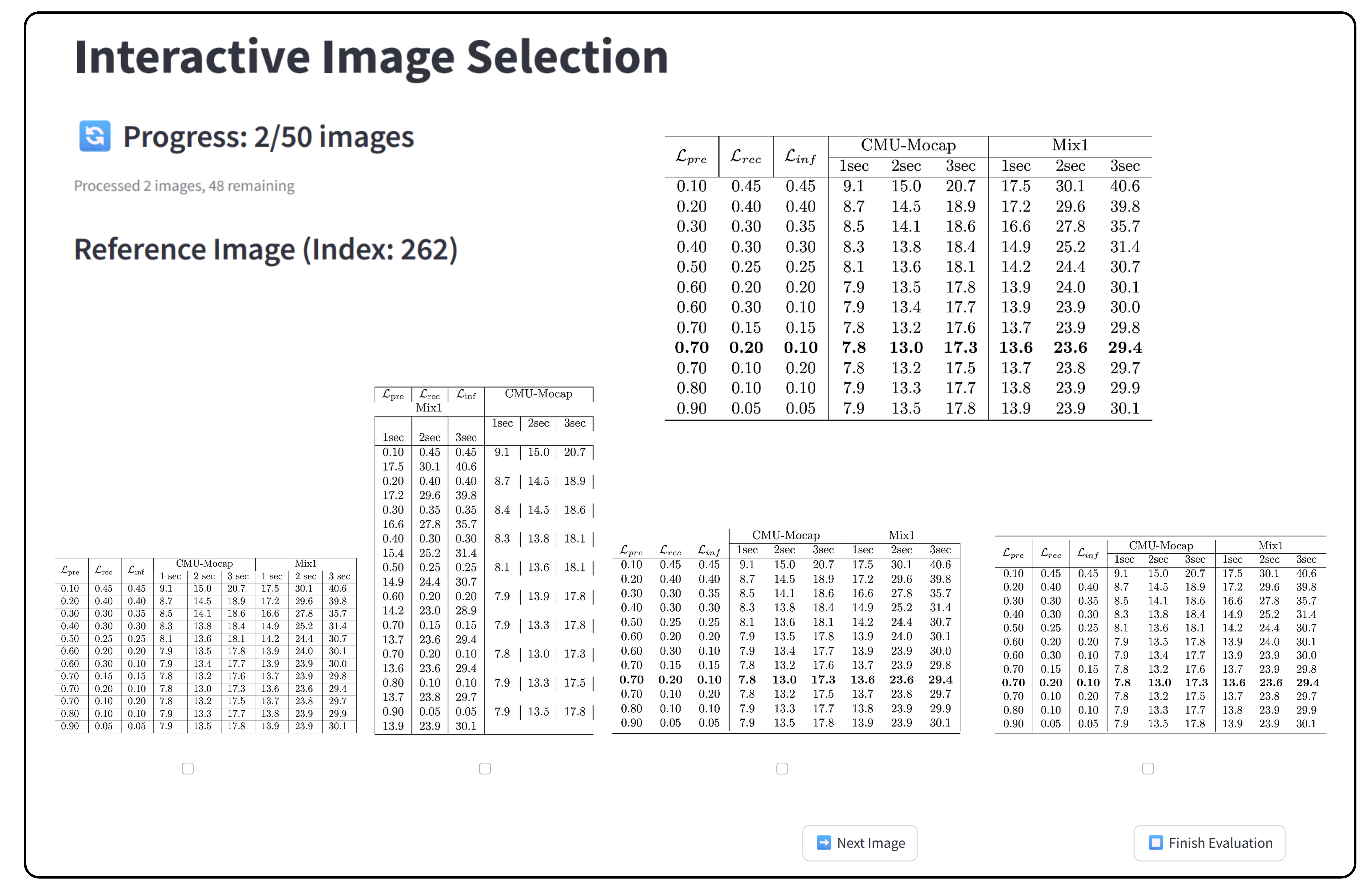}
    \caption{The table image selection page for human evaluation.}
    \label{fig:human_evaluation}
\end{figure}

To evaluate whether the table images rendered from the model-generated LaTeX better align with human preferences, we place the ground truth table image at the top and display the model's predicted table image below it, side by side. The order is randomly shuffled, and the names are hidden. We place the ground truth table image at the top, allowing humans to select one or more images based on subjective similarity. As shown in \Cref{fig:human_evaluation}.


\newpage
\section*{NeurIPS Paper Checklist}

\begin{enumerate}

\item {\bf Claims}
    \item[] Question: Do the main claims made in the abstract and introduction accurately reflect the paper's contributions and scope?
    \item[] Answer: \answerYes{} 
    \item[] Justification: Please refer to the Introduction.

\item {\bf Limitations}
    \item[] Question: Does the paper discuss the limitations of the work performed by the authors?
    \item[] Answer: \answerYes{} 
    \item[] Justification: Please refer to the Limitations.

\item {\bf Theory assumptions and proofs}
    \item[] Question: For each theoretical result, does the paper provide the full set of assumptions and a complete (and correct) proof?
    \item[] Answer: \answerNA{} 
    \item[] Justification: This work does not include a theoretical result.

    \item {\bf Experimental result reproducibility}
    \item[] Question: Does the paper fully disclose all the information needed to reproduce the main experimental results of the paper to the extent that it affects the main claims and/or conclusions of the paper (regardless of whether the code and data are provided or not)?
    \item[] Answer: \answerYes{} 
    \item[] Justification: We have demonstrated all technical details to reproduce the results.

\item {\bf Open access to data and code}
    \item[] Question: Does the paper provide open access to the data and code, with sufficient instructions to faithfully reproduce the main experimental results, as described in supplemental material?
    \item[] Answer: \answerNA{} 
    \item[] Justification: The data and the code would be released after acceptance.

\item {\bf Experimental setting/details}
    \item[] Question: Does the paper specify all the training and test details (e.g., data splits, hyperparameters, how they were chosen, type of optimizer, etc.) necessary to understand the results?
    \item[] Answer: \answerYes{} 
    \item[] Justification: Please refer to the experimental setup.

\item {\bf Experiment statistical significance}
    \item[] Question: Does the paper report error bars suitably and correctly defined or other appropriate information about the statistical significance of the experiments?
    \item[] Answer: \answerNo{} 
    \item[] Justification: Due to resource limitations, we do not repeat one of the experiments. Running a complete experiment takes one week.

\item {\bf Experiments compute resources}
    \item[] Question: For each experiment, does the paper provide sufficient information on the computer resources (type of compute workers, memory, time of execution) needed to reproduce the experiments?
    \item[] Answer: \answerYes{} 
    \item[] Justification: We include the device information in the experimental setup.
    
\item {\bf Code of ethics}
    \item[] Question: Does the research conducted in the paper conform, in every respect, with the NeurIPS Code of Ethics \url{https://neurips.cc/public/EthicsGuidelines}?
    \item[] Answer:  \answerYes{} 
    \item[] Justification: We have reviewed the NeuroIPS Code of Ethics and checked the paper in every respect.

\item {\bf Broader impacts}
    \item[] Question: Does the paper discuss both potential positive societal impacts and negative societal impacts of the work performed?
    \item[] Answer: \answerNo{} 
    \item[] Justification: This work is not related to any private or personal data, and there’s no explicit negative social impacts.
    
\item {\bf Safeguards}
    \item[] Question: Does the paper describe safeguards that have been put in place for responsible release of data or models that have a high risk for misuse (e.g., pretrained language models, image generators, or scraped datasets)?
    \item[] Answer: \answerNA{} 
    \item[] Justification: No such models or datasets are involved.

\item {\bf Licenses for existing assets}
    \item[] Question: Are the creators or original owners of assets (e.g., code, data, models), used in the paper, properly credited and are the license and terms of use explicitly mentioned and properly respected?
    \item[] Answer: \answerYes{} 
    \item[] Justification: We have cited the original papers that offer the original idea or technical details.

\item {\bf New assets}
    \item[] Question: Are new assets introduced in the paper well documented and is the documentation provided alongside the assets?
    \item[] Answer: \answerNo{} 
    \item[] Justification: The paper does not release new assets.

\item {\bf Crowdsourcing and research with human subjects}
    \item[] Question: For crowdsourcing experiments and research with human subjects, does the paper include the full text of instructions given to participants and screenshots, if applicable, as well as details about compensation (if any)? 
    \item[] Answer: \answerYes{} 
    \item[] Justification: We invite 5 graduate students majoring in computer science to participate in the volunteer-based evaluation. The instructions below are provided for each evaluator: at the top of the page, a reference table image is presented, followed by four model-generated table images. You are asked to anonymously select the image that best matches the reference. Preferably, choose only one. If multiple candidates appear equally similar, multiple selections are allowed. In cases where no image is clearly similar, prioritize those with the most similar structural layout.

\item {\bf Institutional review board (IRB) approvals or equivalent for research with human subjects}
    \item[] Question: Does the paper describe potential risks incurred by study participants, whether such risks were disclosed to the subjects, and whether Institutional Review Board (IRB) approvals (or an equivalent approval/review based on the requirements of your country or institution) were obtained?
    \item[] Answer: \answerNA{} 
    \item[] Justification: The paper does not involve crowdsourcing nor research with human subjects.

\item {\bf Declaration of LLM usage}
    \item[] Question: Does the paper describe the usage of LLMs if it is an important, original, or non-standard component of the core methods in this research? Note that if the LLM is used only for writing, editing, or formatting purposes and does not impact the core methodology, scientific rigorousness, or originality of the research, declaration is not required.
    \item[] Answer: \answerYes{} 
    \item[] Justification: We use the pretrained MLLM in the task of table image to LaTeX code generation.

\end{enumerate}

\end{document}